\RequirePackage{fix-cm}
\documentclass[smallextended]{svjour3}

\smartqed  

\usepackage{graphicx}
\usepackage{amsmath}
\usepackage{amssymb}
\usepackage{algorithmic}
\usepackage{algorithm}
\usepackage{subfigure}
\usepackage{array}
\usepackage{tabularx}
\usepackage{multirow}
\usepackage{color}

\newcommand{\bfx}{{\textbf{x}}}
\newcommand{\bfv}{{\textbf{v}}}
\newcommand{\bfu}{{\textbf{u}}}
\newcommand{\bfw}{{\textbf{w}}}

\begin{document}

\title{Sparse learning of maximum likelihood model for optimization of complex loss function}

\author{Ning Zhang
\and
Prathamesh Chandrasekar}

\institute{
N. Zhang\at
Guangzhou Civil Aviation College, Guangzhou 510403, China\\
\email{zhangning115@yahoo.com}\\
\and
P. Chandrasekar
\at
Uttar Pradesh Technical University, Lucknow, Uttar Pradesh 226021, India\\
\email{prathameshchandrasekar@yahoo.com}
}

\date{Received: date / Accepted: date}

\maketitle

\begin{abstract}
Traditional machine learning methods usually minimize a simple loss function to learn a predictive model, and then use a complex performance measure to measure the prediction performance. However, minimizing a simple loss function cannot guarantee that an optimal performance. In this paper, we study the problem of optimizing the complex performance measure directly to obtain a predictive model. We proposed to construct a maximum likelihood model for this problem, and to learn the model parameter, we minimize a complex loss function corresponding to the desired complex performance measure. To optimize the loss function, we approximate the upper bound of the complex loss. We also propose impose the sparsity to the model parameter to obtain a sparse model. An objective is constructed by combining the upper bound of the loss function and the sparsity of the model parameter, and we develop an iterative algorithm to minimize it by using the fast iterative shrinkage-thresholding algorithm framework. The experiments on optimization on three different complex performance measures, including F-score, receiver operating characteristic curve, and recall precision curve break even point, over three real-world applications, aircraft event recognition of civil aviation safety, intrusion detection in wireless mesh networks, and image classification, show the advantages of the proposed method over state-of-the-art methods.
\end{abstract}

\section{Introduction}

Machine learning aims to train a predictive model from a training set of input-out pairs, and then use the model to predict an unknown output from a given test input \cite{liu2015supervised,Zhao20141317,Xu20141303,Liu201516,wang2015representing,Barbosa2015154,Neoh201572}. In this paper, we focus on the machine learning problem of binary pattern classification. In this problem, each input is a feature vector of a data point, and each output is a binary class label of a data point, either positive or negative \cite{He2013793,Tian20141007,Ryu20152254,SanchezValdes2015100,wang2015image,wang2015multiple,wang2014effective}. To learn the predictive model, i.e., the classification model, we usually compare the true class label of data point against the predicted label using a loss function. For example, hinge loss, logistic loss, and squared $\ell_2$ norm loss. By minimizing the loss functions over the training set with regard to the parameter of the classification model, we can obtain a optimal classification model. To evaluate the performance of the model, we apply it to a set of test data points to predict their class labels, and then compare the predicted class labels to their true class labels. This comparison can be conducted by using some multivariate performance measures. For example, prediction accuracy, F-score, area under receiver operating characteristic curve (AUROC) \cite{Jaimeez2015178,Schlattmann201561,Sedgwick2015,Bhattacharya201573}, and precision-recall curve break-even point (RPBEP) \cite{Wen2014421,Ozenne2015855,Villmann201571,Saito2015}. A problem of such machine learning procedure is that in the training process, we optimize a simple loss function, such as hinge loss, but in the test process, we use a different and complex performance measure to evaluate the prediction results. It is obvious that the optimization of the loss function cannot leads to a optimization of the performance measure. For example, in the formulation of support vector machine (SVM), the hinge loss is minimized, but usually in the test procedure, the AUROC is used as a performance measure. However, in many real-world application, the optimization of a specific performance measure is desired. To solve this problem, direct optimization of some complex loss functions corresponding to some desired performance measures are studied. These methods try to a complex loss function in the objective function, and the loss functions is corresponding to the performance measure directly. By minimizing the loss function directly to obtain the predictive model, the desired performance measure can be optimized by the predictive model directly. In this paper, we study this problem, and propose a novel method based on sparse learning and maximum likelihood optimization.

\subsection{Related works}

Some existing works proposed to optimize a complex multivariate loss function are briefly introduced as follows.

\begin{itemize}
\item Joachims proposed to learn a support vector machine to optimize a complex loss function \cite{Joachims2005377}. In the proposed model, the complexity of the predictive model is reduced by minimizing squared $\ell_2$ norm of the model parameter. To minimize the complex loss, its upper bound is approximated and minimized.

\item Mao and Tsang improved the Joachims's work by integrating feature selection to support vector machine for complex loss optimization \cite{mao2013feature}. A weight is assigned to each feature before the predictive model is learned. Moreover, the feature weights and the predictive model parameter is learned jointly in an iterative algorithm.

\item Li et al. proposed a classifier adaptation method to extend Joachims's work \cite{li2013efficient}. The predictive model is a combination of a base classifier and an adaptation function, and the learning of the optimal model is transferred to the learning of the parameter of the adaptation function.

\item {Zhang et al. \cite{zhang2012smoothing} proposed a novel smoothing strategy by using Nesterov's accelerated gradient method to improve the convergence rate of the method proposed by Joachims \cite{Joachims2005377}. This method, according to the results reported in \cite{zhang2012smoothing}, can achieve converges significantly faster than Joachims's method \cite{Joachims2005377}, but it does not scarify generalization ability.}

\end{itemize}

Almost all the existing methods are limited to the support vector machine for multivariate complex loss function. This method uses a linear function to construct the predictive model, and seek both the minimum complexity and loss.

\subsection{Contribution}

In this paper, we propose a novel predictive model to optimize a complex loss function. This model is based on the likelihood of a positive or negative class given a input feature vector of a data point. The likelihood function is constructed based on a Sigmoid function of a linear function. Given a group of data points, we organize them as a data tuple, and the predicted class label tuple is the one that maximize the logistic likelihood of the data tuple. The learning target is to learn a predictive model parameter, so that with the corresponding predicted class label tuple, the complex loss function can be minimized. Moreover, we also hope the model parameter can be as sparse as possible, so that only the useful can be kept in the model. To this end, we construct an objective function, which is composed of two terms. The first term is the $\ell_1$ norm of the parameter to impose the sparsity of the parameter, and the second term is the complex loss function to seek the optimal desired performance measure. The problem is transferred to a minimization problem of the objective function with regard to the parameter. To solve this problem, we first approximate the upper bound of the complex as a logistic function of the parameter, and then optimize it by using the fast iterative shrinkage-thresholding algorithm (FISTA). The novelty of this paper is summarized as follows,

\begin{enumerate}
\item For the first time, we propose to use the maximum likelihood model to construct a predictive model for the optimization of complex losses.
\item We construct a novel optimization problem for the learning of the model parameter by considering the sparsity of the model, and the minimization of the complex loss jointly.
\item We develop a novel iterative algorithm to optimize the proposed minimization problem, and a novel method to approximate the upper bound of the complex loss. The approximation of the upper bound of the complex loss is obtained as a logistic function, and the problem is optimized by a FISTA algorithm.
\end{enumerate}

\subsection{Paper organization}

This paper is organized as follows: In section \ref{sec:method}, we introduce the proposed method, in section \ref{sec:experiment}, we evaluate the proposed method on two real-world applications, and in section \ref{sec:conclusion}, the paper is concluded and some future works are given.

\section{Proposed method}
\label{sec:method}

\subsection{Problem formulation}

Suppose we have a data set of $n$ data points, and we denote them as $\{(\bfx_i,y_i)\}|_{i=1}^n$, where $\bfx_i \in \mathbb{R}^d$ is the $d$-dimensional feature vector of the $i$-th data point, and $y_i\in \{+1,-1\}$ is it corresponding class label. We consider the data points as a data tuple, $\overline{\bfx}=(\bfx_1, \cdots, \bfx_n)$, and their corresponding class labels as a label tuple, $\overline{y}=(y_1, \cdots, y_n)$. Under the framework of complex performance measure optimization, we try to learn a multivariate mapping function  to map the data tuple $\overline{\bfx}$ to a class label tuple $\overline{y}^* = (y_1^*,\cdots,y_n^*)\in \mathcal{Y}$, where $y_i^*\in \{+1,-1\}$ is the predicted label of the $i$-th data point, and $\mathcal{Y}= \{+1,-1\}^n$. To measure the performance of the multivariate mapping function, $\overline{h}(\overline{\bfx})$, we use a predefined complex loss function $\Delta(\overline{y},\overline{y}^*)$ to compare the true class label tuple $\overline{y}$ against the predicted class label tuple $\overline{y}^*$.

To construct the multivariate mapping function $\overline{h}(\overline{\bfx})$, we proposed to apply a linear discriminate function to match the $i$-th data point $\bfx_i$ against the $i$-th class label $y_i'$ in a candidate tuple $\overline{y}'=(y_1', \cdots, y_n')$,

\begin{equation}
\begin{aligned}
f_\bfw(\bfx_i,y_i') = y_i' \bfw^\top \bfx_i,
\end{aligned}
\end{equation}
where $\bfw = [w_1,\cdots,w_d]\in \mathbb{R}^d$ is the parameter vector of the function. And then we apply a Sigmoid function to the response of this function to impose it to a range of $[0,1]$,

\begin{equation}
\begin{aligned}
g(\bfx_i,y_i')
&= \frac{1}{1+ \exp\left (- f(\bfx_i,y_i') \right )} \\
&= \frac{1}{1+ \exp\left (-y_i' \bfw^\top \bfx_i\right )}.
\end{aligned}
\end{equation}
Moreover,

\begin{equation}
\begin{aligned}
g(\bfx_i,+1)
&= \frac{1}{1+ \exp\left (- \bfw^\top \bfx_i\right )}\\
&= \frac{\left ( 1+ \exp\left (- \bfw^\top \bfx_i\right ) \right )
- \exp\left (- \bfw^\top \bfx_i\right ))}{1+ \exp\left (- \bfw^\top \bfx_i\right )}\\
&= 1-\frac{ \exp\left (- \bfw^\top \bfx_i\right ))}{1+ \exp\left (- \bfw^\top \bfx_i\right )}\\
&= 1-\frac{1}{1+ \exp\left (\bfw^\top \bfx_i\right )}\\
&= 1-g(\bfx_i,-1),
\end{aligned}
\end{equation}
thus we can treat $g(\bfx_i,y_i)$ as the conditional probability of $y= y_i'$ given $\bfx = \bfx_i$,

\begin{equation}
\begin{aligned}
Pr(y=y_i'| \bfx = \bfx_i) = g(\bfx_i,y_i').
\end{aligned}
\end{equation}
We also assume that the data points in the tuple $\overline{\bfx}$ are conditional independent from each other, and thus the conditional probability of $\overline{y} = \overline{y}'$ given the $ \overline{\bfx}$ is

\begin{equation}
\begin{aligned}
Pr(\overline{y}=\overline{y}'| \overline{\bfx} )
& = \prod_{i=1}^n Pr(y=y_i'| \bfx = \bfx_i)\\
&= \prod_{i=1}^n \frac{1}{1+ \exp\left (-y_i' \bfw^\top \bfx_i\right )}.
\end{aligned}
\end{equation}
To constructed the complex mapping function, we map the data tuple to the class tuple $\overline{y}^*$ which can give the maximum log-likelihood,

\begin{equation}
\label{equ:y_star}
\begin{aligned}
y^*\leftarrow \overline{h}(\overline{\bfx})
& = \underset{\overline{y}'\in \mathcal{Y}}{\arg\max}~ \log \left ( Pr(\overline{y}=\overline{y}'| \overline{\bfx} ) \right )\\
&= \underset{\overline{y}'\in \mathcal{Y}}{\arg\max} ~ \log \left (  \prod_{i=1}^n \frac{1}{1+ \exp\left (-y_i' \bfw^\top \bfx_i\right )} \right ).
\end{aligned}
\end{equation}
In this way, we seek the maximum likelihood estimator of the class label tuple as the mapping result for a data tuple.

To learn the parameter of the linear discriminative function, $\bfw$, so that the complex loss function $\Delta(\overline{y},\overline{y}^*)$ can be minimized, we consider the following problems,

\begin{itemize}
\item \textbf{Encouraging sparsity of $\bfw$}:
We assume that in a feature vector a data point, only a few features are useful, while most of the remaining features are useless. Thus we need to conduct a feature selection procedure to remove the useless features and keep the useful features, so that we can obtain a parse feature vector. In our method, instead of seeking sparsity of the feature vectors, we seek the sparsity of the parameter vector $\bfw$. With a sparse $\bfw$, we can also control the sparsity of the feature effective to the prediction results. To encourage the sparsity of $\bfw$, we use the $\ell_1$ norm of $\bfw$ to present its sparsity, and minimize the $\ell_1$,

\begin{equation}
\label{equ:obj1}
\begin{aligned}
\min_{\bfw}
&\left \{ \frac{1}{2} \left \| \bfw \right \|_1
 = \frac{1}{2} \sum_{j=1}^d |w_j|
\right .\\
& \left.
=\frac{1}{2} \sum_{j=1}^d \frac{w_j^2}{|w_j|}
=\frac{1}{2} \bfw^\top diag \left (\frac{1}{|w_1|},\cdots, \frac{1}{|w_d|} \right ) \bfw
\right.\\
& \left . =
\frac{1}{2} \bfw^\top \Lambda \bfw
\vphantom{\frac{1}{2} \sum_{1}}
\right \},
\end{aligned}
\end{equation}
where $diag \left (\frac{1}{|w_1|},\cdots, \frac{1}{|w_d|} \right ) \in \mathbb{R}^{d\times d}$ is a diagonal matrix with its diagonal elements as $\frac{1}{|w_1|},\cdots, \frac{1}{|w_d|}$, and

\begin{equation}
\label{equ:Lambda}
\begin{aligned}
\Lambda = diag \left (\frac{1}{|w_1|},\cdots, \frac{1}{|w_d|} \right )
\end{aligned}
\end{equation}
When the $\ell_1$ norm of $\bfw$ is minimized, most elements of $\bfw$ will shrink to zeros, and leads a sparse $\bfw$.

\item \textbf{Minimizing complex performance lose $\Delta(\overline{y},\overline{y}^*)$}: Given the predicted label tuple $\overline{y}^*$, we can measure the prediction performance by comparing it against the true label tuple $\overline{y}$ by using a complex performance measure. To obtain an optimal mapping function,  we minimize a corresponding complex loss of a complex performance measure, $\Delta(\overline{y},\overline{y}^*)$,

\begin{equation}
\label{equ:objDelta}
\begin{aligned}
\min_{\bfw}~
&\Delta(\overline{y},\overline{y}^*)
\end{aligned}
\end{equation}
Due to its complexity, we minimize its upper boundary instead of itself. We have the following Theorem to define the  upper boundary of $\Delta(\overline{y},\overline{y}^*)$.

\begin{description}
\item
\textbf{Theorem 1}: $\Delta(\overline{y},\overline{y}^*)$ satisfies

\begin{equation}
\label{equ:upper}
\begin{aligned}
\Delta(\overline{y},\overline{y}^*)\leq &
\max_{\overline{y}'\in \mathcal{Y}}
\left \{
\log \left (  \prod_{i=1}^n \frac{1}{1+ \exp\left (-y_i' \bfw^\top \bfx_i\right )} \right )
-
\log \left (  \prod_{i=1}^n \frac{1}{1+ \exp\left (-y_i \bfw^\top \bfx_i\right )} \right )
\right .\\
& \left .
+ \Delta(\overline{y},\overline{y}')
\vphantom{\sum_1 \frac{1}{2}}
\right \}\\
& =
\left \{
\log \left (  \prod_{i=1}^n \frac{1}{1+ \exp\left (-y_i'' \bfw^\top \bfx_i\right )} \right )
-
\log \left (  \prod_{i=1}^n \frac{1}{1+ \exp\left (-y_i \bfw^\top \bfx_i\right )} \right )
\right .\\
& \left .
+ \Delta(\overline{y},\overline{y}'')
\vphantom{\sum_1 \frac{1}{2}}
\right \},
\end{aligned}
\end{equation}
where $\overline{y}' = (y_1',\cdots,y_n')$, and $\overline{y}'' = (y_1',\cdots,y_n')$,

\begin{equation}
\label{equ:upper1}
\begin{aligned}
\overline{y}'' =
&
\arg\max_{\overline{y}'\in \mathcal{Y}}
\left \{
\log \left (  \prod_{i=1}^n \frac{1}{1+ \exp\left (-y_i' \bfw^\top \bfx_i\right )} \right )
-
\log \left (  \prod_{i=1}^n \frac{1}{1+ \exp\left (-y_i \bfw^\top \bfx_i\right )} \right )
\right .\\
& \left .
+ \Delta(\overline{y},\overline{y}')
\vphantom{\sum_1 \frac{1}{2}}
\right \}
\end{aligned}
\end{equation}

The proof of this Theorem is found in Appendix section.

\end{description}

After we have the upper bound of the loss function, we minimize it instead of $\Delta(\overline{y},\overline{y}^*)$ to obtain the mapping function parameter, $\bfw$,

\begin{equation}
\label{equ:obj2}
\begin{aligned}
\min_{\bfw}
&
\left \{
\log \left (  \prod_{i=1}^n \frac{1}{1+ \exp\left (-y_i'' \bfw^\top \bfx_i\right )} \right )
-
\log \left (  \prod_{i=1}^n \frac{1}{1+ \exp\left (-y_i \bfw^\top \bfx_i\right )} \right )
\right .\\
&
+ \Delta(\overline{y},\overline{y}'')
\vphantom{\sum_1 \frac{1}{2}}
\\
&
\left .
=\sum_{i=1}^n \log \left (
\frac{1+\exp(-y_i\bfw^\top \bfx_i)}{1+\exp(-y_i''\bfw^\top \bfx_i)}
\right )+\Delta(\overline{y},\overline{y}'')
\right \}.
\end{aligned}
\end{equation}
Please note that $\overline{y}''$ is also a function of $\bfw$.

\end{itemize}

The overall optimization problem is obtained by combining the problems in (\ref{equ:obj1}) and (\ref{equ:obj2}),

\begin{equation}
\label{equ:obj3}
\begin{aligned}
\min_{\bfw}
~&
\left \{
f(\bfw)=
\frac{1}{2} \bfw^\top \Lambda \bfw+ C
\left [
\sum_{i=1}^n \log \left (
\frac{1+\exp(-y_i\bfw^\top \bfx_i)}{1+\exp(-y_i''\bfw^\top \bfx_i)}
\right )+\Delta(\overline{y},\overline{y}'')
\right ]
\right \}
\end{aligned}
\end{equation}
where $C$ is a tradeoff parameter. Please not that in this objective, both $\Lambda$ and $\overline{y}''$ are functions of $\bfw$. In the first term of the objective, we impose the sparsity of the $\bfw$, and in the second term, we minimize the upper bound of $\Delta(\overline{y},\overline{y}^*)$.

\subsection{Optimization}

To solve the problem of (\ref{equ:obj3}), we try to employ the FISTA algorithm with constant step-size to minimize the objective $f(\bfw)$.  This algorithm is an iterative algorithm, and in each iteration, we first update a search point according to a previous solution of the parameter vector, and then update the next parameter vector based on the search point. The basic procedures are summarized as the two following steps:

\begin{enumerate}
\item \textbf{Search point step}: In this step, we assume the previous solution of $\bfw$ is $\bfw_{pre}$, and seek a search point $\bfv\in \mathbb{R}^d$ based on $\bfw$ is $\bfw_{pre}$ and a stepsize $L$.
\item \textbf{Weighting factor step}: In this step, we assume we have a weighting factor of previous iteration, $\tau_{pre}$, and we update it to a new weighting factor $\tau_{cur}$.
\item \textbf{Solution update step}: In this step, we update the new solution of the variable according to the search point. The updated solution is a weighted version of the previous search points, weighted by the weighting factors.
\end{enumerate}

In the follows, we will discuss how to implement these three steps.

\subsubsection{Search point step}

In this step, when we want to minimize an objective function $f(\bfw)$ with regard to a variable vector $\bfw$ with a step-size $L$ and a previous solution $\bfw_{pre}$, we seek a search point $\bfu^*$ as follows,

\begin{equation}
\label{equ:search}
\begin{aligned}
\bfu^* = \arg\min_{\bfu}
~&
\left \{
\frac{L}{2} \left \| \bfu - \left ( \bfw_{pre} - \frac{1}{L} \nabla f(\bfw_{pre})\right )\right \|_2^2
\right \},
\end{aligned}
\end{equation}
where $\nabla f(\bfw)$ is the gradient function of $f(\bfw)$. Due to the complexity of function $f(\bfw)$, the close form of gradient function $\nabla f(\bfw)$ is difficult to obtain. Thus instead of seeking gradient function directly, we seek the sub-gradient of this function. This end, we use the EM algorithm strategy. In each iteration, we first fix $\bfw$ as $\bfw_{pre}$, and calculate $\Lambda$ according to (\ref{equ:Lambda}), and $y_i''|_{i=1}^n$ according to (\ref{equ:upper1}). Then we fix $\Lambda$ and $y_i''|_{i=1}^n$ and seek the sub-gradient $\nabla f(\bfw)$,

\begin{equation}
\label{equ:subgradient}
\begin{aligned}
\nabla f(\bfw)=
\Lambda\bfw+ C
\sum_{i=1}^n  \left (
\frac{y_i'' \bfx_i \exp(-y_i''\bfw^\top \bfx_i)}{1+\exp(-y_i''\bfw^\top \bfx_i)}
- \frac{y_i \bfx_i \exp(-y_i\bfw^\top \bfx_i)}{1+\exp(-y_i\bfw^\top \bfx_i)}
\right ).
\end{aligned}
\end{equation}
After we have the sub-gradient function $\nabla f(\bfw)$, we substitute it to (\ref{equ:search}), and we have

\begin{equation}
\label{equ:search1}
\begin{aligned}
\bfu^* = \arg\min_{\bfu}
~&
\left \{
\frac{L}{2} \left \| \bfu - \left ( \bfw_{pre} - \frac{1}{L} \nabla f(\bfw_{pre})\right )\right \|_2^2
\right \}\\
= \arg\min_{\bfu}
~&
\left \{
\frac{L}{2} \left \| \bfu - \left [ \bfw_{pre} - \frac{1}{L}
\left (
\Lambda\bfw_{pre}+ C
\sum_{i=1}^n  \left (
\frac{y_i'' \bfx_i \exp(-y_i''\bfw_{pre}^\top \bfx_i)}{1+\exp(-y_i''\bfw_{pre}^\top \bfx_i)}
\right . \right . \right . \right . \right . \\
&
\left. \left. \left. \left. \left.
- \frac{y_i \bfx_i \exp(-y_i\bfw_{pre}^\top \bfx_i)}{1+\exp(-y_i\bfw_{pre}^\top \bfx_i)}
\right )\vphantom{\sum_1 \frac{1}{1}}
\right )
\right ]\right \|_2^2
\right \}\\
= \arg\min_{\bfu}
~&
\left \{
\frac{L}{2} \left \| \bfu - \left [ \left (I - \frac{1}{L} \Lambda \right )\bfw_{pre} - \frac{C}{L}
\sum_{i=1}^n  \left (
\frac{y_i'' \bfx_i \exp(-y_i''\bfw_{pre}^\top \bfx_i)}{1+\exp(-y_i''\bfw_{pre}^\top \bfx_i)}
\right . \right . \right . \right . \\
&
\left. \left. \left. \left.
- \frac{y_i \bfx_i \exp(-y_i\bfw_{pre}^\top \bfx_i)}{1+\exp(-y_i\bfw_{pre}^\top \bfx_i)}
\right )\vphantom{\sum_1 \frac{1}{1}}
\right ]\right \|_2^2 = g(\bfu)
\right \}.
\end{aligned}
\end{equation}
To solve this problem, we set the gradient function of the objective function $g(\bfu)$ to zero,

\begin{equation}
\label{equ:gu}
\begin{aligned}
& \nabla g(\bfu) =
L \left \{ \bfu - \left [ \left (I - \frac{1}{L} \Lambda \right )\bfw_{pre} - \frac{C}{L}
\sum_{i=1}^n  \left (
\frac{y_i'' \bfx_i \exp(-y_i''\bfw_{pre}^\top \bfx_i)}{1+\exp(-y_i''\bfw_{pre}^\top \bfx_i)}
\right . \right . \right .\\
&
\left. \left. \left.
- \frac{y_i \bfx_i \exp(-y_i\bfw_{pre}^\top \bfx_i)}{1+\exp(-y_i\bfw_{pre}^\top \bfx_i)}
\right )\vphantom{\sum_1 \frac{1}{1}}
\right ]\right \} = 0\\
& \Rightarrow
\bfu^* =  \left (I - \frac{1}{L} \Lambda \right )\bfw_{pre} - \frac{C}{L}
\sum_{i=1}^n  \left (
\frac{y_i'' \bfx_i \exp(-y_i''\bfw_{pre}^\top \bfx_i)}{1+\exp(-y_i''\bfw_{pre}^\top \bfx_i)}
\right . \\
&
\left.
- \frac{y_i \bfx_i \exp(-y_i\bfw_{pre}^\top \bfx_i)}{1+\exp(-y_i\bfw_{pre}^\top \bfx_i)}
\right )\vphantom{\sum_1 \frac{1}{1}}.
\end{aligned}
\end{equation}
In this way, we obtain the search point $\bfu^*$.

\subsubsection{Weighting factor step}

We assume that weighting factor of previous iteration is $\tau_{pre}$, we can obtain the weighting factor of current iteration, $\tau_{cur}$, as follows,

\begin{equation}
\label{equ:factor}
\begin{aligned}
\tau_{cur}=
\frac{1+\sqrt{1+4{\tau_{pre}}^2}}{2}.
\end{aligned}
\end{equation}

\subsubsection{Solution update step}

After we have the search point of this current iteration, $\bfu^*$, the search point of previous iteration, ${\bfu^*}_{pre}$, and the weighting factor of this iteration and previous iteration, $\tau_{cur}$ and $\tau_{pre}$, we can have the following update procedure for the solution of this iteration,

\begin{equation}
\label{equ:updatew}
\begin{aligned}
\bfw_{cur}
&=\bfu^* + \left ( \frac{\tau_{pre} - 1}{\tau_{cur}}\right ) \left ( \bfu^* - \bfu^*_{pre}\right )\\
&=\left ( \frac{\tau_{cur} + \tau_{pre} - 1}{\tau_{cur}}\right ) \bfu^* -  \left ( \frac{\tau_{pre} - 1}{\tau_{cur}}\right ) \bfu^*_{pre}.
\end{aligned}
\end{equation}
In this equation, we can see that the updated solution of $\bfw_{cur}$ is a weighted version of the current search point, $\bfu^*$, and the previous search point, $\bfu^*_{pre}$.

\subsection{Iterative algorithm}

With the optimization in the previous section, we summarize the iterative algorithm to optimize the problem in (\ref{equ:obj3}). The iterative algorithm is given in Algorithm 1.

\begin{enumerate}
\item[] \textbf{Algorithm 1: FISTA with constant stepsize to optimize (\ref{equ:obj3})}

\item \textbf{Input}: $L$, a constant step-size;

\item \textbf{Step 0}: Take $\bfw_1$ = $\bfu_0$, $\tau_1 = 1$.

\item \textbf{Step $k$ ($k \geq 1$)}:

\begin{enumerate}

\item Update $\Lambda_k$ according to (\ref{equ:Lambda}) by fixing $\bfw = \bfw_k$.

\item Update ${y_i''}_k|_{i=1}^n$ according to (\ref{equ:upper1}) by fixing $\bfw = \bfw_k$.

\item Update $\bfu_k$ according to (\ref{equ:gu}) by fixing $\bfw_{pre} = \bfw_k$, $\Lambda =\Lambda_k$, and ${y_i''} = {y_i''}_k, i=1,\cdots, n$.

\item Updating $\tau_k$ according to (\ref{equ:factor}) by fixing $\tau_{k-1} = \tau_{pre}$.

\item Updating $\bfw_k$ according to (\ref{equ:updatew}) by fixing $\bfu^* = \bfu_k$, $\bfu^*_{pre} = \bfu_{k-1}$, $\tau_{cur} = \tau_{k}$, and $\tau_{pre} = \tau_{k-1}$.
\end{enumerate}
\item \textbf{Output}: $\bfw_k$
\end{enumerate}

In this algorithm, we can see that in each iteration, we first update $\Lambda$ and ${y_i''}|_{i=1}^n$, and then use them to update the search point. With the search point and a updated weighting factor, we update the mapping function parameter vector, $\bfw$. This algorithm is called learning of sparse maximum likelihood model (SMLM).

\subsection{{Scaling up to big data based on Hadoop}}

{In this section, we discuss how to fit the proposed algorithm to big data set. We assume that the number of the training data points, $n$, is extremely large. One single machine is not able to store the entire data set, the data set is split to $m$ subsets, and stored in $m$ different clusters. The clusters are managed by a big data platform, Hadoop \cite{Feller201580,Yin201558,Kim20151,Shi20152300}. Hadoop is a software of distributed data management and processing. Given a large data set, it split it to subsets, and store them in different clusters. To process the data and obtain a final output, it uses a Map-Reduce framework \cite{Irudayasamy2015221,Shanoda2015PDC80,Maitrey2015703,Csar201569}. This framework requires a Map program and a Reduce program from the users. The Hadoop software deliver the Map program to each cluster and uses it to process the subset to produce some median results, and then use the Reduce program to combine the median results to produce the final outputs. Using the Map-Reduce the framework, by defining our own Map and Reduce functions, we can implement the critical steps in Algorithm 1. For example, in the sub-step (c) of step $k$, we need to calculate $\bfu_k$ from (\ref{equ:gu}). In this step, the most time consuming step is to calculate the summation of a function over all the data points,}

{\begin{equation}
\begin{aligned}
output
&=
\sum_{i=1}^n  \left (
\frac{y_i'' \bfx_i \exp(-y_i''\bfw_{pre}^\top \bfx_i)}{1+\exp(-y_i''\bfw_{pre}^\top \bfx_i)}
- \frac{y_i \bfx_i \exp(-y_i\bfw_{pre}^\top \bfx_i)}{1+\exp(-y_i\bfw_{pre}^\top \bfx_i)}
\right )\\
&=\sum_{i=1}^n function(\bfx_i, y_i, y_i''),
\end{aligned}
\end{equation}
where $function(\bfx_i, y_i,  y_i'') = \frac{y_i'' \bfx_i \exp(-y_i''\bfw_{pre}^\top \bfx_i)}{1+\exp(-y_i''\bfw_{pre}^\top \bfx_i)}
- \frac{y_i \bfx_i \exp(-y_i\bfw_{pre}^\top \bfx_i)}{1+\exp(-y_i\bfw_{pre}^\top \bfx_i)}$ is the function applied to each data point. Since the entire data set is split to $m$ subsets, $\mathcal{X}_m|_{j=1}^m$, we can design a Map function to calculate the summation over each subset, and then design a Reduce function to combine the to obtain the final output. The Map and Reduce functions are as follows.}

{\begin{enumerate}
\item[] \textbf{Map function applied to the $j$-th subset}
\item \textbf{Input}: Data points of the $j$-th subset, $\{(\bfx_i, y_i,y_i'')\}|_{i:\bfx_i\in \mathcal{X}_j}$.
\item \textbf{Input}: Previous parameter, $\bfw_{pre}$.
\item \textbf{Initialize}: $Output_j = 0$.
\item \textbf{For~$i:\bfx_i\in \mathcal{X}_j$}
\begin{enumerate}
\item $Output_j = Output_j + function(\bfx_i, y_i, y_i'')$;
\end{enumerate}
\item \textbf{Endfor}
\item \textbf{Output}: $Output_j$
\end{enumerate}}

{\begin{enumerate}
\item[] \textbf{Reduce function to calculate the final output}
\item \textbf{Input}: Median outputs of $m$ Map functions, $Output_j|_{j=1}^m$.
\item \textbf{Initialize}: $Output = 0$.
\item \textbf{For~$j=1,\cdots, m$}
\begin{enumerate}
\item $Output = Output + Output_j$;
\end{enumerate}
\item \textbf{Endfor}
\item \textbf{Output}: $Output$
\end{enumerate}}

\section{Experiment}
\label{sec:experiment}

In this section, we evaluate the proposed SMLM for the optimization of complex loss function. Three different applications are considered, which are aircraft event recognition, intrusion detection in wireless mesh networks, and image classification.

\subsection{Aircraft event recognition}

Recognizing aircraft event of aircraft landing is an important problem in the area of civil aviation safety research. This procedure provides important information for fault diagnosis and structure maintenance of aircraft \cite{Wang2014201}. Given a landing condition, we want to predict if it is normal and abnormal. To this end, we extract some features, and use them to predict the aircraft event of normal or abnormal. In this experiment, we evaluate the proposed algorithm in this application, and use it as a model for the prediction of aircraft event recognition.

\subsubsection{Data set}

In this experiment, we collect a data set of 160 data points. Each data point is a landing condition, and we describe the landing condition by five features, including vertical acceleration, vertical speed, lateral acceleration, roll angle, and pitch rate. The data points are classified to two classes, normal class and abnormal. The normal class is treated as positive class, while the abnormal class is treated as negative class. The number of positive data points is 108, and the number of negative data points is 52.

\subsubsection{Experiment setup}

In this experiment, we use the 10-fold cross validation. The data set is split into 10 folds randomly, and each fold contains 16 data points. Each fold is used as a test set in turn, and the remaining 9 folds are combined and used as training set. The proposed model is training over the training set, and then used to predict the class labels of the testing data points in the test set. The prediction results are evaluated by a performance measurement. This performance measurement is used to compare the true class labels of the test data points against the predicted class labels. In the training procedure, a complex loss function corresponding to the performance measurement is minimized.

In our experiments, we consider three performance measurements, which are F-score, area under receiver operating characteristic curve (AUROC), and precision-recall curve break-even point (RPBEP). To define these performance measures, we first need to define the following items,

\begin{itemize}
\item true positive (TP), the number of correctly predicted positive data points,
\item true negative (TN), the number of correctly predicted negative data points,
\item false positive (FP), the number of negative data points wrongly predicted to positive data points, and
\item false negative (FN), the number of positive data points wrongly predicted to negative data points.
\end{itemize}
With these measures, we can define F-score as follows,

\begin{equation}
\begin{aligned}
F = \frac{2\times TP}{2\times TP+FP+FN}.
\end{aligned}
\end{equation}
Moreover, we can also define true positive rate (TPR) and the false positive rate (FPR) as follows,

\begin{equation}
\begin{aligned}
TPR = \frac{TP}{TP+FN}, ~FPR = \frac{FP}{FP+TN}.
\end{aligned}
\end{equation}
With different thresholds we can have different pair TPR and FPR. By plotting TPR against FPR values, we can have a curve of receiver operating characteristic (ROC). The area under this curve is obtained as AUROC. The recall and precision are defined as follows,

\begin{equation}
\begin{aligned}
recall = \frac{TP}{TP+FN}, ~precision = \frac{TP}{TP+FP}.
\end{aligned}
\end{equation}
With different thresholds, we can also have different pair of recall and precision values. We can obtain a recall-precision (RP) curve, by plotting different precision values against recall values. RPBEP is the value of the point of the RP curve where recall and precision are equal to each other.

\subsubsection{Experiment result}

\begin{figure}[!h]
\centering
\includegraphics[width=0.6\textwidth]{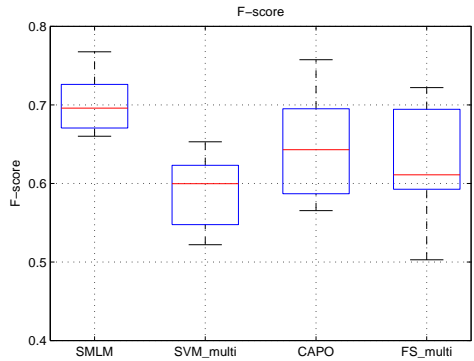}
\caption{Boxplots of F-score of compared method on aircraft event recognition problem.}
\label{fig:aircraft_F}
\end{figure}

\begin{figure}[!h]
\centering
\includegraphics[width=0.6\textwidth]{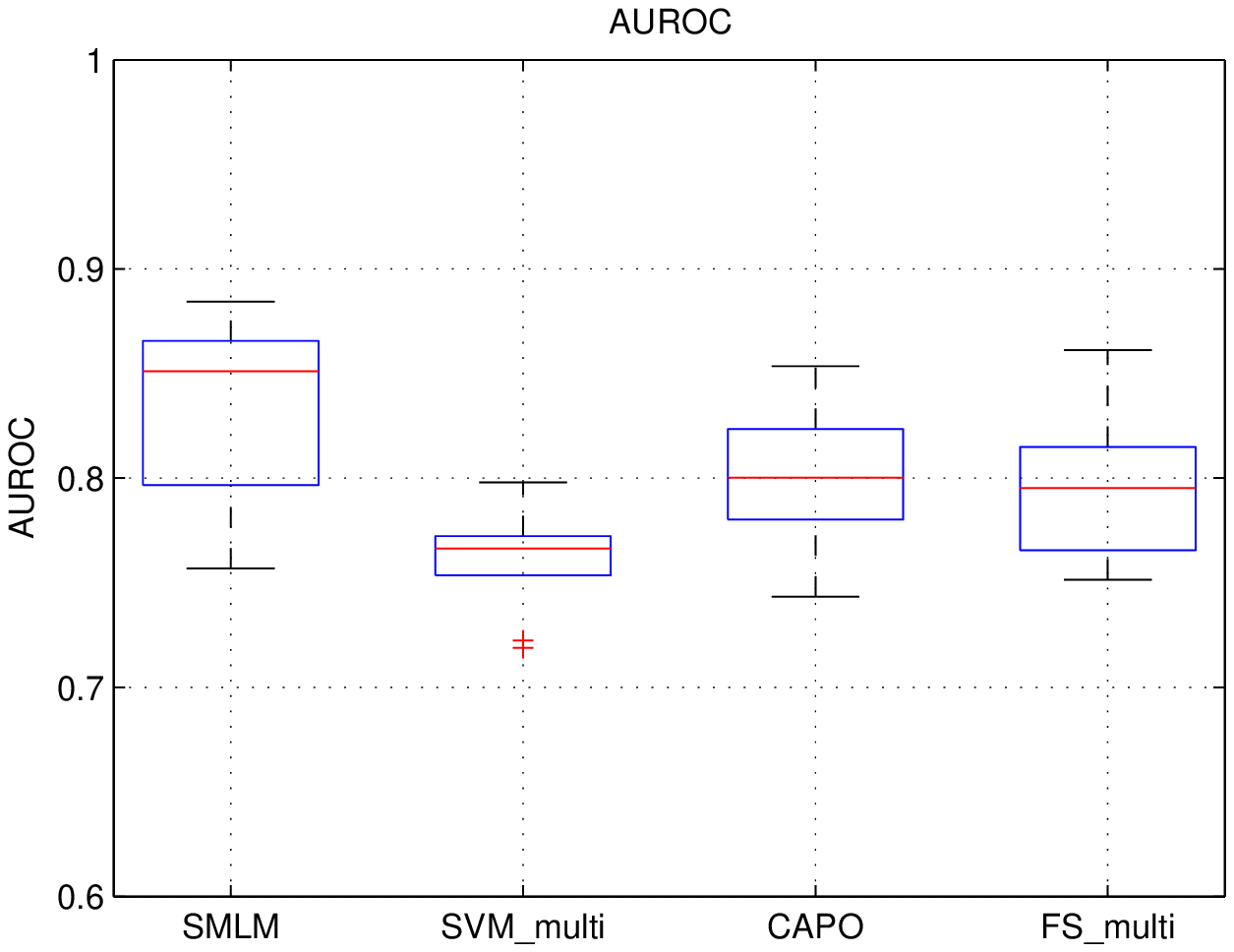}
\caption{Boxplots of AUROC of compared method on aircraft event recognition problem.}
\label{fig:aircraft_AUROC}
\end{figure}

\begin{figure}[!h]
\centering
\includegraphics[width=0.6\textwidth]{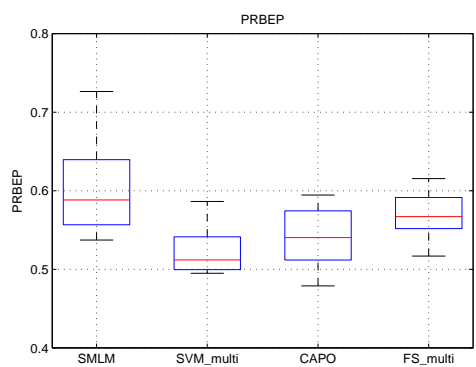}
\caption{Boxplots of PRBEP of compared method on aircraft event recognition problem.}
\label{fig:aircraft_PRBEP}
\end{figure}

We compare the proposed algorithm, SMLM, against several state-of-the-art complex loss optimization methods, including support vector machine for multivariate performance optimization (SVM$_{multi}$) \cite{joachims2005support}, classifier adaptation for multivariate performance optimization (CAPO) \cite{li2013efficient}, and features selection for multivariate performance optimization (FS$_{multi}$) \cite{mao2013feature}. The boxplots of the optimized F-scores of 10-fold cross validation of different algorithms on the aircraft event recognition problem are given in Fig. \ref{fig:aircraft_F}, these of optimized AUROC are given in Fig. \ref{fig:aircraft_AUROC}, and these of the optimized PRBEP are given Fig. \ref{fig:aircraft_PRBEP}. From these figures, we can see that the proposed method, SMLM, outperforms the compared algorithms on three different optimized performances. For example, in Fig. \ref{fig:aircraft_PRBEP}, we can see that the boxplot of PRBEP of SMLM is significantly higher than other methods, the median value is almost 0.6, while that of other methods are much lower than 0.6. In Fig. \ref{fig:aircraft_AUROC}, we can also have similar observation, the overall AUROC values optimized by SMLM is much higher than these of other methods. A reason for this outperforming is that our method seeks the maximum likelihood and sparsity of the model simultaneously.

\subsection{Intrusion detection in wireless mesh networks}

Wireless mesh network (WMN) is a new generation technology of wireless networks, and it has been used in many different applications. However, due to its openness in wireless communication, it is vulnerable to intrusions, thus it is extremely important to detect intrusion in WMN. Given an attack record, the problem of intrusion detection is to classify it to one of the following classes, denial service attacks, detect attacks, obtain root privileges and remote attack unauthorized access
attacks. In this paper, we use the proposed method, SMLM, for the problem intrusion detection,

\subsubsection{Data set}

In this experiment, we use the KDD CPU1999 data set. In this data set, contains 4,0000 attack records, and for each class, there are 1,0000 records. For each record, we first preprocess the record, and then convert the features into digital signature as the new features.

\subsubsection{Experiment setup}

In this experiment, we also use the 10-fold cross validation, and we also use the F-score, AUROC, and RPBEP performance measures.

\subsubsection{Experiment result}

\begin{figure}[!h]
\centering
\includegraphics[width=0.6\textwidth]{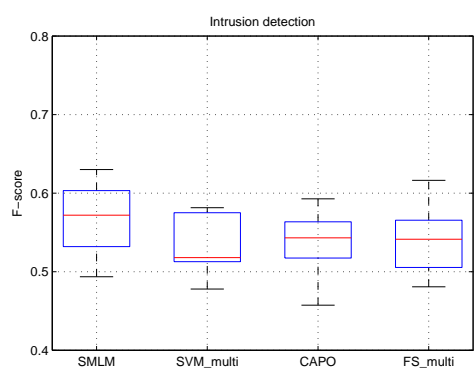}
\caption{Boxplots of F-score of compared method on intrusion detection problem.}
\label{fig:intrusion_F}
\end{figure}

\begin{figure}[!h]
\centering
\includegraphics[width=0.6\textwidth]{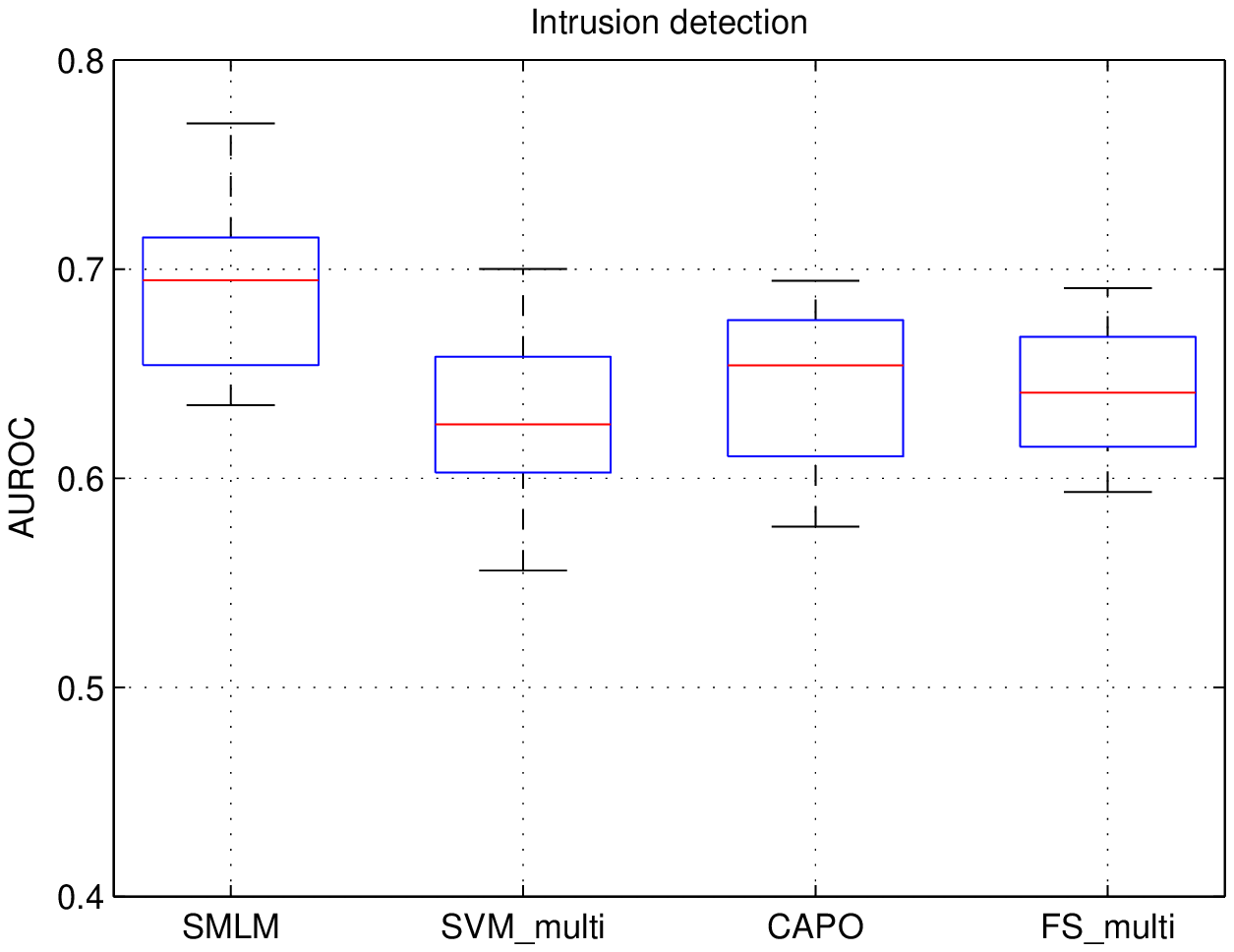}
\caption{Boxplots of AUROC of compared method on aircraft event recognition problem.}
\label{fig:intrusion_AUROC}
\end{figure}

\begin{figure}[!h]
\centering
\includegraphics[width=0.6\textwidth]{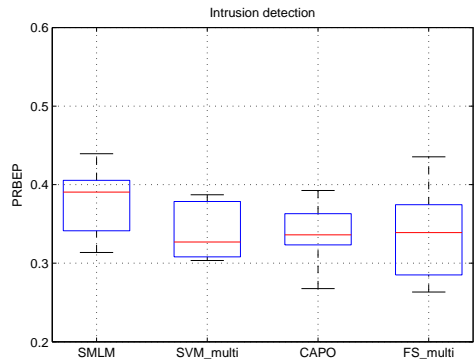}
\caption{Boxplots of PRBEP of compared method on aircraft event recognition problem.}
\label{fig:intrusion_PRBEP}
\end{figure}

The boxplots of the optimized F-scores of 10-fold cross validation are given in Fig. \ref{fig:intrusion_F}, the boxplots of AUROC are given in Fig. \ref{fig:intrusion_AUROC}, and the boxplots of PRBEP are given in Fig. \ref{fig:intrusion_PRBEP}. Similar to the results on aircraft event recognition problem, the outperforming of the proposed algorithm, SMLM, over other methods are also significant. This is a strong evidence of the advantages of sparse learning and maximum likelihood.

{\subsection{ImageNet image classification}}

{In the third experiment, we use a large image set to test the performance of the proposed algorithm with big data.}

{\subsubsection{Data set}}

{In this experiment, we use a large data set, ImageNet \cite{Krizhevsky20121097}. This data set contains over 15 million images, and the images belong to 22,000 classes. These images are from web pages, and are labeled by people manually. The entire data set are split into three subsets, which are one training set, one validation set, and on testing set. The training set contains 1.2 million images, the validation set contains 50,000 images, and the testing set contains 150,000 images. To represent each image, we use the bag-of-features method. Local SIFT features are extracted from each image, and quantized to a histogram. The features can be downloaded directly from http://image-net.org/download-features.}

{\subsubsection{Experiment setup}}

{In this experiment, we do not use the 10-fold cross validation, but use the given training/ validation/ testing set splitting. We first perform the proposed algorithm to the training set to learn the classifier, then use the validation set to justify the optimal tradeoff parameters, finally test the classifier over the testing set. The performances of F-score, AUROC, and RPBEP are considered in this experiment. To handle the multi-classification problem, we have a binary classification problem for each class, and in this problem, the considered class is a positive class, while the combination of all other classes is a negative class.}

{\subsubsection{Experiment results}}

\begin{figure}
  \centering
  \includegraphics[width=0.6\textwidth]{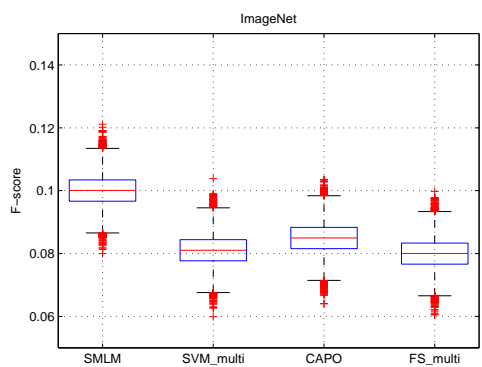}\\
  \caption{Boxplots of F-score of compared method on ImageNet image classifiation problem.}
  \label{fig:imageFscore}
\end{figure}

\begin{figure}
  \centering
  \includegraphics[width=0.6\textwidth]{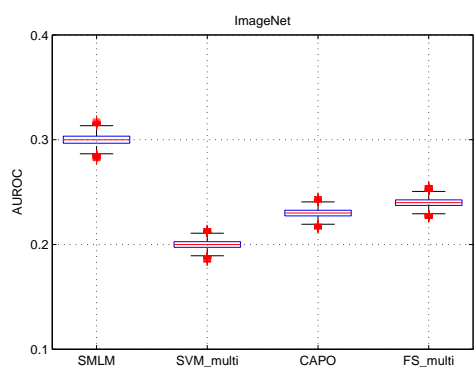}\\
  \caption{Boxplots of AUROC of compared method on ImageNet image classifiation problem.}
  \label{fig:imageAUROC}
\end{figure}

\begin{figure}
  \centering
  \includegraphics[width=0.6\textwidth]{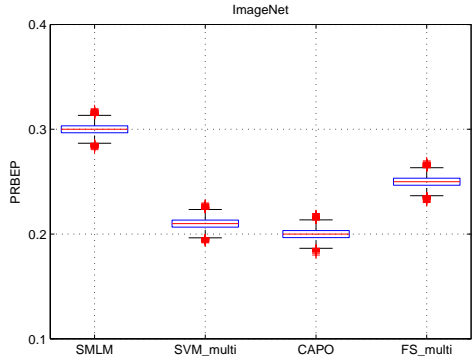}\\
  \caption{Boxplots of PRBEP of compared method on ImageNet image classifiation problem.}
  \label{fig:imagePRBEP}
\end{figure}

{The Boxplots of the optimized F-score,  AUROC, and RPBEP of different classes are given in Fig. \ref{fig:imageFscore}, Fig. \ref{fig:imageAUROC}, and Fig. \ref{fig:imagePRBEP}. From these figures, we clearly see that the proposed algorithm outperforms the competing methods. This is another strong evidence of the effectiveness of the SMLM algorithm. Moreover, it also shows that the proposed algorithm also works well over the big data.}

\subsection{{Running time}}

\begin{figure}
  \centering
  \includegraphics[width=0.6\textwidth]{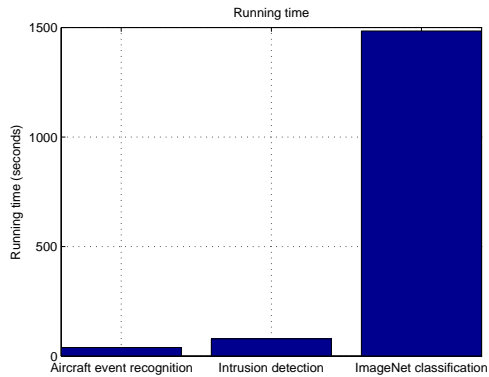}\\
  \caption{Running time of SMLM algorithm on three experiments.}
  \label{fig:runningtime}
\end{figure}

{The running time of the proposed algorithm on the three used data sets are given in Fig. \ref{fig:runningtime}. It can be observed from this figure that the first two experiments do not consume much time, while the third large scale data set based experiment costs a lot of time. This is natural, because in each iteration of the algorithm, we have a function for each data point, and a summation over the responses of this function.}

\section{Conclusion}
\label{sec:conclusion}

In this paper, we investigate the problem of optimization of complex corresponding to a complex multivariate performance measure. We propose a novel predicative model to solve this problem. This model is based on the maximum likelihood of a class label tuple given a input data tuple. To solve the model parameter, we propose an optimization problem based on the approximation of the upper bound of the loss function and the sparsity of the model. Moreover, an iterative algorithm is developed to solve it. Experiments on two real-world applications show its advantages over state-of-the-art. In the future, we will extend the proposed algorithm to different applications, such as computational biology \cite{wang2014computational,zhou2014biomarker,liu2013structure,peng2015modeling} and multimedia information processing \cite{wang2015representing,wang2015supervised,wang2015multiple,wang2015image}.


\section*{{Appendix}}

\begin{description}

\item
{\textbf{Proof of Theorem 1}: According to (\ref{equ:y_star}), we have}

{\begin{equation}
\label{equ:y_star1}
\begin{aligned}
\overline{y}^* &= \underset{\overline{y}'\in \mathcal{Y}}{\arg\max} ~ \log \left (  \prod_{i=1}^n \frac{1}{1+ \exp\left (-y_i' \bfw^\top \bfx_i\right )} \right ),\\
\Rightarrow
&
\log \left (  \prod_{i=1}^n \frac{1}{1+ \exp\left (-y_i^* \bfw^\top \bfx_i\right )} \right )
\geq
\log \left (  \prod_{i=1}^n \frac{1}{1+ \exp\left (-y_i' \bfw^\top \bfx_i\right )} \right ), \forall \overline{y}' \in \mathcal{Y},\\
\Rightarrow
&
\log \left (  \prod_{i=1}^n \frac{1}{1+ \exp\left (-y_i^* \bfw^\top \bfx_i\right )} \right )
\geq
\log \left (  \prod_{i=1}^n \frac{1}{1+ \exp\left (-y_i \bfw^\top \bfx_i\right )} \right ), \\
\Rightarrow
&
\log \left (  \prod_{i=1}^n \frac{1}{1+ \exp\left (-y_i^* \bfw^\top \bfx_i\right )} \right )
-
\log \left (  \prod_{i=1}^n \frac{1}{1+ \exp\left (-y_i \bfw^\top \bfx_i\right )} \right )
\geq
0, \\
\Rightarrow
&
\log \left (  \prod_{i=1}^n \frac{1}{1+ \exp\left (-y_i^* \bfw^\top \bfx_i\right )} \right )
-
\log \left (  \prod_{i=1}^n \frac{1}{1+ \exp\left (-y_i \bfw^\top \bfx_i\right )} \right )\\
&
+\Delta(\overline{y},\overline{y}^*)
\geq
\Delta(\overline{y},\overline{y}^*), \\
\Rightarrow
&
\max_{\overline{y}'\in \mathcal{Y}}\left \{
\log \left (  \prod_{i=1}^n \frac{1}{1+ \exp\left (-y_i' \bfw^\top \bfx_i\right )} \right )
-
\log \left (  \prod_{i=1}^n \frac{1}{1+ \exp\left (-y_i \bfw^\top \bfx_i\right )} \right )
\right .\\
&
\left .
+\Delta(\overline{y},\overline{y}')
\vphantom{\sum_1}
\right \}
\geq
\Delta(\overline{y},\overline{y}^*),
\end{aligned}
\end{equation}
thus we have (\ref{equ:upper}).}

\end{description}


\end{document}